# A Hackathon for Classical Tibetan

Orna Almogi[1], Lena Dankin[2*], Nachum Dershowitz[2,3], Lior Wolf[2]

[1]Universität Hamburg, Germany

[2]Tel Aviv University, Israel

[3]Institut d'Études Avancées de Paris, France

[*]Corresponding author: Lena Dankin, lenadank@tau.ac.il

**Abstract**
We describe the course of a hackathon dedicated to the development of linguistic tools for Tibetan Buddhist studies. Over a period of five days, a group of seventeen scholars, scientists, and students developed and compared algorithms for intertextual alignment and text classification, along with some basic language tools, including a stemmer and word segmenter.

## I  INTRODUCTION

In February 2016, a group of four Tibetologists (from the University of Hamburg), one digital humanities scholar (from Europe), and twelve computer scientists (from Israel and Europe) got together in Kibbutz Lotan in the Arava region of Israel with the stated goal of developing algorithmic methods for advancing Tibetan Buddhist textual studies. Participants were either recruited by the organizers or responded to an announcement on several mailing lists. See Figure 1. Most of the computer scientists had background in machine learning, and a few of them also had experience with natural language processing (NLP) research, but without any prior experience with Tibetan texts. The computer scientist organizers were quite familiar with programming workshops and contests and thought that the challenges presented by Tibetan texts would pose an ideal opportunity to explore the hackathon format. The hackathon is a short and intense event where computer scientists collaborate to develop software. For that purpose, it was essential to recruit as many software developers as possible. Some of the recruited students had participated in other hackathons. The plan was to have a focused hacking event, with specific goals to work towards, goals that had been provided by the Tibetan scholars.

The six-hour drive down from Tel Aviv (including a stop to admire desert flora) afforded an opportunity for everyone to get to know each other. The back seats of the van were piled high with computer equipment and the kibbutz was to provide the necessary fast internet connection. The isolation of the kibbutz created an intense working environment and encouraged long hours; the stark natural beauty of the location contributed to a shared sense of tranquility of purpose.

Several hackathons have been conducted for the purpose of the development of tools for digital humanities, before and after ours. In June 2015, the eTRAP team organized a hackathon for text reuses (http://www.etrap.eu/tutorials/2015-goettingen/). Twenty-three participants from fifteen different institutes worked on the detection of textual reuses across data in different languages and from different genres, using the TRACER tool [Büchler, 2013; Büchler et al., 2014]. More recently, in May 2017, another hackathon took place in Helsinki, which brought together historians, linguists, psychologists, and computer scientists to work on four different tasks, including the analysis of the written media with the political elite (https://www.helsinki.fi/en/helsinki-centre-for-digital-humanities/helsinki-digital-humanities-hackathon-2017-dhh17). In November 2017, The National Library of Israel hosted a 24-hour hackathon dedicated to the goal of developing state-of-the-art tools and applications for national cultural treasures using a iiif server for their large collection of images (https://hackathon.nli.org.il/index-en.html).

The two main tasks that confronted our group that week (February 14-18) were (1) to develop algorithms for finding intertextual parallels that are only approximately the same, and (2) to experiment with algorithmic classification methods for identifying authorship and style. In both cases, the concern was centered on language issues specific to Tibetan.

After a quick lesson in Tibetan, the Buddhist canon, and modern Tibetan encoding conventions for the benefit of the less knowledgeable, the group split into four loose teams, devoted to the following goals: (A) dataset preparation; (B) language tool development; (C) intertextual alignment; and (D) text classification. We describe each of these efforts in turn in the sections that follow. Each team consisted of a few computer scientists, chosen based on the individual background, experience, and interests, plus a Tibetan scholar who provided annotated data sets and analyzed results. Twice a day we held synchronization round-ups, where each team briefed everyone about their progress, discussed their next steps, and raised problems they stumbled across.

## II    PRELIMINARIES

Tibetan is a monosyllabic language (Tibetan morphemes normally consist of one syllable) belonging to the Tibeto-Burman branch of the Sino-Tibetan family. The language is ergative, with a plethora of (usually monosyllabic) grammatical particles, which are often omitted. Occasionally, the same syllable can be written using one of several orthographic variations, for example, *sogs* and *stsogs*. In the case of verbs, the syllable has various inflectional forms that are often homophones, a fact that can result in variants in reading due to scribal errors or lack of standardization. An example of such inflectional forms is *sgrub, bsgrubs, bsgrub, sgrubs* (present, past, future and imperative, respectively), all of which are homophones. The intransitive form of the verb offers even more inflectional forms that yield homophones with their transitive counterpart, *'grub* and *grub* (present/future and past, respectively). See [Beyer 1992] for details about the language.

The Tibetan Buddhist canon consists of two parts: the *Kangyur (bKa' 'gyur)*, which commonly comprises 108 volumes containing what is believed by tradition to be the Word of the Buddha, texts that were mostly translated directly from the Sanskrit original (with some from other languages and others indirectly via Chinese); and the *Tengyur (bsTan 'gyur)*, commonly comprising about 210 volumes consisting of canonical commentaries, treatises, and various kinds of manuals that were written in the seventh to thirteenth centuries and likewise mostly translated from Sanskrit, with some works from other languages and a few originally written in Tibetan. Overall, this corpus contains 77 million occurrences (tokens) of 81,000 different syllable types. The average transcribed syllable length is 3.5 and the average number of syllables in a single document is 18201.

## III  HACKATHON TASKS

### A. Dataset preparation

A prerequisite for the main goals of the hackathon was data with which to work, that is, texts to compare and classify. For this, we took Tibetan Buddhist texts obtained from various sources. These included the Tibetan Buddhist canon in digital form (we used a modified form of the ACIP files of the *Kangyur* and *Tengyur* provided by Paul Hackett of Columbia University) and several sets of autochthonous Tibetan Buddhist texts of various authors (compiled by Eric Werner of Universität Hamburg).

In addition, it was necessary to prepare test suites with manually prepared "gold standard" answers, so that the performance of algorithms for finding parallel passages and for classifying texts could be measured. The passages were selected from various sources, particularly from (a) two doxographical texts (*'grub mtha'*), the *gZhung lugs rnam 'byed* by Phywa pa Chos kni sengge (1109–1169) and the *'Grub mtha' mdzod* by Klong chen pa Dri med 'od zer (1308–1364), the latter including "borrowed" passages from the former [Werner, 2014], and (b) Rong zom Chos kyi bzang po's (11th c.) collected writings, which features numerous cases of parallel passages.

These Tibetan works were provided in textual form, transcribed according to the Wylie convention [Wylie, 1959]. In this system, Tibetan is transliterated into Latin characters without diacritics; thus various Tibetan letters are represented by two or three Latin consonants. The decision to work with transliterated texts was made partly because they were the ones available at the time, but also because the computer scientists didn't understand Tibetan script, so this transliteration made it possible for them to progress quickly without the need to acquire a new alphabet. The texts had to be "cleaned" by removing sigla and by standardizing punctuation.

### B. Language tools

Since syllables having the same base form may take many different surface forms, stemming is a crucial stage in almost every text-processing task one would like to perform in Tibetan, as for many other languages. So, to support present and future analysis of Tibetan texts, developing a stemmer was one of the first orders of business.

Usually, in Indo-European and Semitic languages, stemming is performed on the word level. However, in Tibetan, in which multisyllabic words are not separated by spaces or other marks, a syllable-based stemming mechanism is required even in order to segment the text into lexical items. Stemming is not the same as (grammatical) lemmatization, and the stemming process can result in a stem that is not itself a lexical entry in a dictionary. Moreover, unlike Indo-European languages, stemming of Tibetan is mostly relevant to verbs and verbal nouns (which are common in the language). Despite being inaccurate in some cases, stemming (for Tibetan, as for other languages) can improve tasks such as word segmentation and the detection of intertextual parallels [Klein et al., 2014]. Even for Tibetan words consisting of more than one syllable, stemming each "substantial" syllable (i.e. excluding grammatical particles) makes sense since all the inflections are embedded at the syllable level. For instance, the words *brtag dbyad* (analysis) and *brtags dpyad* (analyzed) are stemmed to *rtog dpyod* (to analyze, analysis).

The stemmer we developed is a rule-based application that works in the following manner: first, the syllable is divided into a sequence of Tibetan letters. This stage is required because the Wylie transliteration scheme represents some Tibetan letters by more than one character (e.g. *zh, tsh*). There is, fortunately, no ambiguity in the process of segmentation into Tibetan letters. By design, the transliteration ensures that whenever a sequence of two or three characters represents a single letter, it cannot also be interpreted in context as a sequence of distinct Tibetan letters.

For the analysis of the Tibetan syllable we used an octuple (8-component) scheme: Each Tibetan syllable should contain one core letter and one vowel. Other positions (subscript, superscript, coda, prescript, postscript, and appended particle) are not obligatory. Each position contains a single letter, except for that of the appended particle, which can be any of six syllables. The "stem" of a syllable is defined by us as consisting of the core letter or stacked letter (which, in turn, consists of the core letter and a superscript or a subscript, or both), the vowel (syllabic contractions contain two vowels at most), and the coda (if extant). Syllables can be considered stemmically identical if these are consistent, despite additions or omissions of a prescript and/or a postscript. The final stage of the stemming is normalization, since there are groups of Tibetan letters that can be replaced one with another without changing the basic meaning of the syllable (in inflectional forms). Since the goal is to group all syllables that are ultimately stemmically identical into one and the same stem, we normalized all tuples according to an elaborate set of rules.

The stemmer, as described, extracts the information encoded in each Wylie transliterated syllable and makes it explicit. An important task, given two syllables, is to evaluate their stemmic similarity. Some substitutions can be considered silent or synonymous; others change the meaning completely; and there is a continuous spectrum in between. Metric learning algorithms were used to assess the relative importance of each substitution.

Another important language task is word segmentation, that is, grouping syllables into words (lexical units). Since no spaces or special characters are used to mark word boundaries, the reader has to rely on language models to detect the word boundaries. As opposed to the stemming task, we had recourse to an annotated corpus for the segmentation task, that is, a word-segmented corpus, with which it was possible to train a supervised model. The training data that was used, consisting of 37,000 sentences, was obtained from the Tibetan in Digital Communication project (http://larkpie.net/tibetancorpus).

The approach taken at the hackathon was based on a flavor of recurrent neural networks (RNNs) called "long short-term memory" (LSTM) [Hochreiter & Schmidhuber, 1997]. LSTMs have been used in the past for word segmentation of Chinese text [Chen et al., 2015]. The tuple representation of syllables was used for this purpose; see details in [Almogi et al., 2016]. Several LSTM setups were compared; the best configuration yielded an F1 score of 0.95. In addition, a more traditional algorithm, the conditional random field (CRF), was applied to the data, yielding a lower F1 score of 0.89. This technique was previously applied on Tibetan script in [Liu et al., 2011].

It bears noting that our efforts to train a word2vec model [Mikolov et al., 2013] to represent Tibetan syllables did not result in a solid representation, in the sense that pairs of vectors with high (cosine) similarity did not usually represent synonyms. For that reason, the vector representation that was developed for the stemmer was also essential for the word segmentation task.

Both the stemmer and word segmenter have been made publicly available and can be accessed from http://www.cs.tau.ac.il/~nachumd/Tools.html. Additional details may be found in [Almogi et al., 2016].

**C. Intertextual alignment**

The primary goal of the hackathon was to develop and compare tools for finding parallel passages between Tibetan texts that are the result of either acknowledged citations (with or without attributions) or borrowing (i.e. with no acknowledgement whatsoever). Generally, for determining the history of composition or relative chronology of a text, passages need not match precisely. That is, in addition to the fact that orthographical differences or omission/addition of grammatical particles are of no great significance, it is often the case that cited or borrowed passages are not necessarily reproduced verbatim, but are often slightly paraphrased or shortened, or both. For determining the identity of persons involved in the composition of the text and its transmission—that is, the author, translator, scribe, or editor— the precision of the match is of greater significance, and even variation in orthography or omission/addition of grammatical particles may be relevant. In this regard, however, textual scholars take into consideration that texts were often copied and edited and that through these processes changes could have been introduced into the text, either deliberately—particularly in terms of standardization of orthography and verb inflection, employment of particles, and even substitutions of terminology in cases of archaism—or unintentionally.

Broadly speaking, there are two cases of interest: (a) an approximate alignment of what could be considered to be exactly the same text, that is, an alignment that allows variants that are considered accidental or non-substantial (that is, variations regarding omission/addition or different forms of the same grammatical particles, orthography, inflectional forms in the case of verbs, archaism vs. standardization, and the like), and (b) an approximate alignment of passages that contained the same text but in modified form of some sort, that is, an alignment that allows substantial variants in addition to the non-substantial ones (omission/addition of a substantial syllable, replacement of a substantial syllable by a completely different one, omission/addition of a string of syllables, occurrence of the same syllables in a different order, and the like). To address the problem of substantial variants that could occur also when a (more or less) exact citation or borrowing was intended, that is, such that have been intentionally introduced by either the author himself or by the scribes and editors during the process of transmission, or such that have been unintentionally crept in during the processes of composition and copying, a limited number of substantial variants must be admitted as well.

Three algorithms competed with one another on this task during the hackathon.

1. One algorithm was TRACER [Büchler, 2013; Büchler et al., 2014], based on the "bag of words" representation method. TRACER is a general text reuse detection algorithm with a seven-level architecture. Each step is configurable and can be optimized to specific text reuse tasks and corpora. The steps are preprocessing, featuring, selection, scoring, and post-processing. This approach is called feature-based linking, where only text-reuse units with shared features are compared, as opposed to the comparison of the full text of passages, all against all. All passages are compared by comparing the words they contain, ignoring word order.

2. Another method was based on *Agents for Actors (AfA)* [Küster, 2013], a "digital humanities framework for distributed microservices for text analysis". AfA was originally developed for the purpose of identifying allusions to Shakespearean passages in transcriptions of dialogues in films (hence "actors" in its name). This algorithm compares passages both on the letter and the word level, and therefore catches variations at the orthographic and formulation levels, respectively. While its primary use is to identify references and allusions in texts, in the hackathon, the algorithm was tested to see how well it can also serve to identify parallel passages for very different types of texts in an unrelated language.

3. The third approach was based on an adaptation of the method of [Barsky et al., 2008], designed for matching DNA subsequences, to our problem, as described in [Klein et al., 2014]. This algorithm looks for "all against all approximate matches" (within some given threshold of difference between passages) by rephrasing the problem as finding maximal paths in a matching graph. That method was modified during the hackathon to work with syllable stems as the basic building block, rather than the individual character level used before. This change improved both the run time and the quality of the results. Since, on average, a syllable has 4 characters, the speedup was two orders of magnitude. As for the results, p@10 ("precision at ten", the fraction of the top ten results that are of relevance) increased from 0.67 to 1, and p@20 increased from 0.37 to 0.63. The improvement were due to the fact that with character-wise alignment syllables can share many letters but have no semantic similarity; see [Labenski et al., 2016; Labenski, 2016].

An infrastructure subteam, in addition to keeping everything up and running, parallelized the implementation of the third algorithm to run on a Sparc cluster of computers, located at Tel Aviv University. This is necessary for the ultimate goal, considering the large size of the corpus. The idea is simple: divide the texts into overlapping chunks; then run the original algorithm on all chunks in parallel; finally, piece all the results together.

All three algorithms were tested on a test set that was designed during the hackathon. The two doxological texts mentioned above and known to contain many shared passages were chosen, and 24 pairs of parallel passages were manually annotated. Out of the 24 pairs, the TRACER algorithm retrieved 13 pairs, the AFA algorithm retrieved 12 pairs, and the APBT algorithm retrieved 16.

By finding cited or borrowed passages within the corpora of Indo-Tibetan (i.e. translated) and Tibetan (i.e. autochthonous) Buddhist literature, several research questions can be better addressed:

- determining the history of composition of individual texts;

- determining relative chronology of groups of texts;

- determining the intellectual scholarly milieu in which the texts emerged; and

- determining the intellectual history behind the texts (viz. terminology and concepts).

After identifying parallel passages, one can assess the frequencies of letter/syllable/word replacements in the aligned passages of selected texts or text groups. This can serve to help answer further research questions like: determining editorial policies and processes, such as standardization of orthography, standardization of employment of grammatical particles (i.e. according to the so-called *sandhi* rules); and identifying processes of "revisions" of translated texts.

## D. Text classification

The second major task that was addressed at the hackathon was the question of author profiling. While the question as to what extent the issue of authorship can be addressed in the case of translated texts is yet to be looked into carefully, some general research questions related to authorship fall under the purview of machine classification. These include the following: (a) distinguishing between translated texts and autochthonous texts; (b) identifying the period in which a text was composed, viz. Old Tibetan (7–11th c.), Classical Tibetan I (11–14th c.), or Classical Tibetan II (15–20th c.); (c) determining whether a translated canonical work belongs to the early period of translation (*snga 'gyur*) or the new period (*phyi 'gyur*); (d) in the case of autochthonous literature, differentiating between the so-called "revealed" texts (texts that are portrayed as having been transmitted supernaturally) versus "composed" texts; and (e) identifying an author's intellectual milieu (e.g. affiliation with a particular school of thought).

A series of experiments were performed on scriptures and treatises, early and late, translated and autochthonous texts. We tried several methods, including bag-of-word features and a perceptron classifier with stochastic gradient descent with features similar to [Volansky et al., 2015], mainly: mean syllable length; mean sentence length; frequency of verbal prefixes and function words; frequency of foreign (Sanskrit) words; and type-to-token ratio. For authorship detection, we first used an automatic word segmenter and then used n-gram frequency and bag-of-words as features. Such a method was shown to be useful in [Koppel et al., 2008]. We didn't advance further in this task, due to a shortage of time.

Both parts of the canon were employed as training data to determine features that are peculiar for the *Kangyur*, the corpus containing scriptures, on the one hand, and the *Tengyur*, the corpus containing treatises, commentaries, manuals and the like, on the other. Numerous autochthonous texts, including the entire collected writings of Rong zom Chos kyi bzang po, the entire collected writings of Shākya mchog ldan (1428–1507), several works by Sa kya paṇḍi ta Kun dga' rgyal mtshan (1182–1251), and several texts by Tsong kha pa Blo bzang grags pa (1357–1419) were tested against the translated canonical texts in order to determine features of translated versus autochthonous works. In addition, selected individual texts were tested. For example, Sa skya paṇḍi ta's *Tshad ma rigs gter* was compared with Dharmakīrti's (7th c.) *Pramāṇavarttika* in Tibetan translation, which enabled a comparison of autochthonous versus translated work on similar topics. The *Mañjuśrīnāmasaṅgīti* commentary ascribed to Rong zom pa (and at the same time included in the *Tengyur* as an Indian work in Tibetan translation) was compared with the canon in its entirety, as was the *Tengyur* alone with other works by Rong zom pa and additional autochthonous works, which provided a comparison of works whose origin has been considered doubtful with translated and autochthonous literature. The classification results are undergoing analysis by the Tibetan scholars.

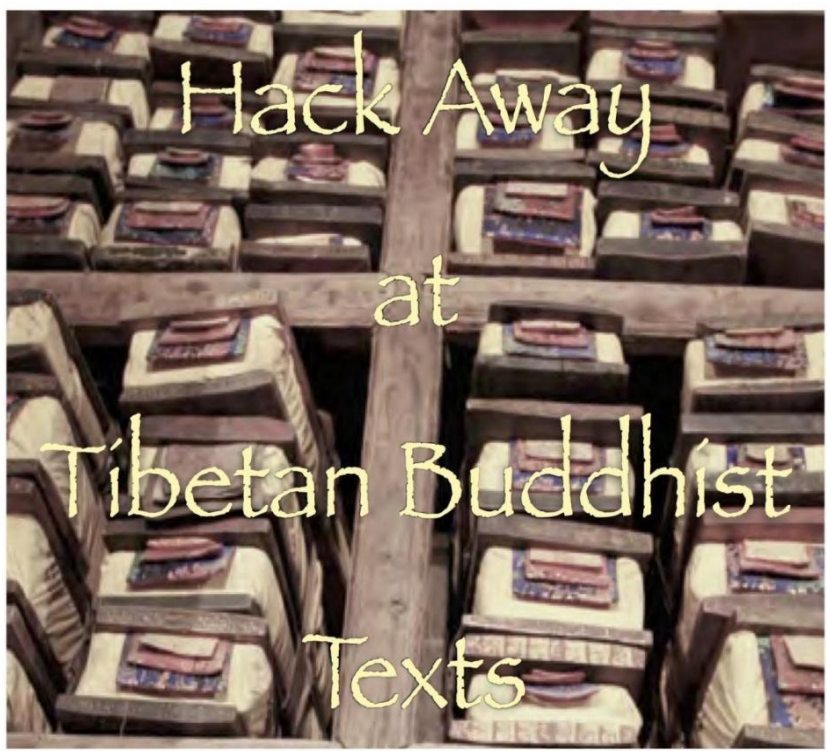

Figure 1. Poster announcement of the hackathon.

IV   CONCLUSION

The intense hackathon format proved to be quite exhilarating. Towards evening, each group reported on the day's accomplishments and vicissitudes. No single task was actually brought to completion on site, but the saplings were planted, and the ideas and prototype tools have continued to grow and develop in the ensuing weeks.

Based on our experience, we would recommend such a hackathon format for other well-defined interdisciplinary efforts in the computational humanities. It pays to come well-prepared to the event with clear goals and clean test data. And it is crucial to allocate resources for bringing the products and results of the hackathon to a stable and useful state after the event.

As a matter of fact, the authors held a second hackathon one year later (February 2017) on a kibbutz in the Galilee, again for the development of tools for Tibetan Buddhist texts, but this time concentrating on manuscripts and computer-vision aspects.

**Acknowledgements**

We thank the staff at Kibbutz Lotan and all the hackathon participants (listed below). This research was supported in part by a grant (#I-145-101.3-2013) from the German-Israeli Foundation for Scientific Research and Development, and by the Khyentse Center for Tibetan Buddhist Textual Scholarship, Universität Hamburg, thanks to a grant by the Khyentse Foundation. N.D.'s research benefitted from a fellowship at the Paris Institute for Advanced Studies (France), with the financial support of the French state, managed by the French National Research Agency's "Investissements d'avenir" program (ANR-11-LABX-0027-01 Labex RFIEA+).

*Hackathon participants:* Orna Almogi, Kfir Bar, Marco Büchler, Lena Dankin, Nachum Dershowitz, Daniel Hershcovich, Yair Hoffman, Marc W. Küster, Daniel Labenski, Peter Naftaliev, Dimitri Pauls, Elad Shaked, Nadav Steiner, Lior Uzan, Dorji Wangchuk, Eric Werner, and Lior Wolf.

*Participating institutions*: Tel Aviv University (School of Computer Science); Universität Hamburg (Khyentse Center for Tibetan Buddhist Textual Scholarship, Department for Indian and Tibetan Studies); Georg-August-Universität Göttingen (Göttingen Centre for Digital Humanities).